\documentclass[10pt,twocolumn,letterpaper]{article}

\usepackage{cvpr}
\usepackage{times}
\usepackage{epsfig}
\usepackage{graphicx}
\usepackage{amsmath}
\usepackage{amssymb}

\usepackage[breaklinks=true,bookmarks=false]{hyperref}
\let\OLDthebibliography\thebibliography
\renewcommand\thebibliography[1]{
  \OLDthebibliography{#1}
  \setlength{\parskip}{0pt}
  \setlength{\itemsep}{1pt plus 0.3ex}
}
\cvprfinalcopy 

\usepackage{listings,lstautogobble}

\def\cvprPaperID{****} 

\begin{document}

\title{Detecting and counting tiny faces}

\author{Alexandre Attia\\
ENS Paris-Saclay\\
{\tt\small alexandre.attia@ens-paris-saclay.fr}
\and
Sharone Dayan\\
ENS Paris-Saclay\\
{\tt\small sharone.dayan@ens-paris-saclay.fr}
}

\maketitle

\begin{abstract}
   Finding Tiny Faces [1] - released at CVPR 2017 - proposes a novel approach to find small objects in an image. Our contribution consists in deeply understanding the choices of the paper together with applying and extending a similar method to a real world subject which is the counting of people in a public demonstration.
\end{abstract}

\section{Introduction}
The paper[1] deals with finding small objects (particularly faces in our case) in an image, based on scale-specific detectors by using features defined over single (deep) feature hierarchy : Scale Invariance, Image resolution, Contextual reasoning.
The algorithm is based on "foveal" descriptors, i.e blurring the peripheral image to encode and give just enough information about the context, mimicking the human vision. The subject is still an open challenge and we would like to experiment this approach to different applications. Thus, after presenting the method and the influence of its parameters, we will concentrate on a real-world application which is counting people in a public demonstration. The last part will focus on possible extensions of this work. Our code (using Python and TensorFlow) is available at \href{https://github.com/alexattia/ExtendedTinyFaces}{github.com/alexattia/ExtendedTinyFaces}

\section{Method}
\subsection{Contextual Reasoning}
It is very difficult - if not impossible - to detect very small faces, even for a human user. Thus, we need to establish how best to encode context. For the context modelling, the paper makes use of a fixed-size (291px) receptive field. It then defines templates over the "hypercolumn" features (vector of activations of all units for each pixel) extracted from multiple layers of a deep model which are effective "foveal" descriptors. This technique allows to capture both high-resolution detail and coarse low-resolution cues across large receptive field.

\subsection{Global Architecture}
\begin{figure*}
\centering
  \includegraphics[width=\textwidth]{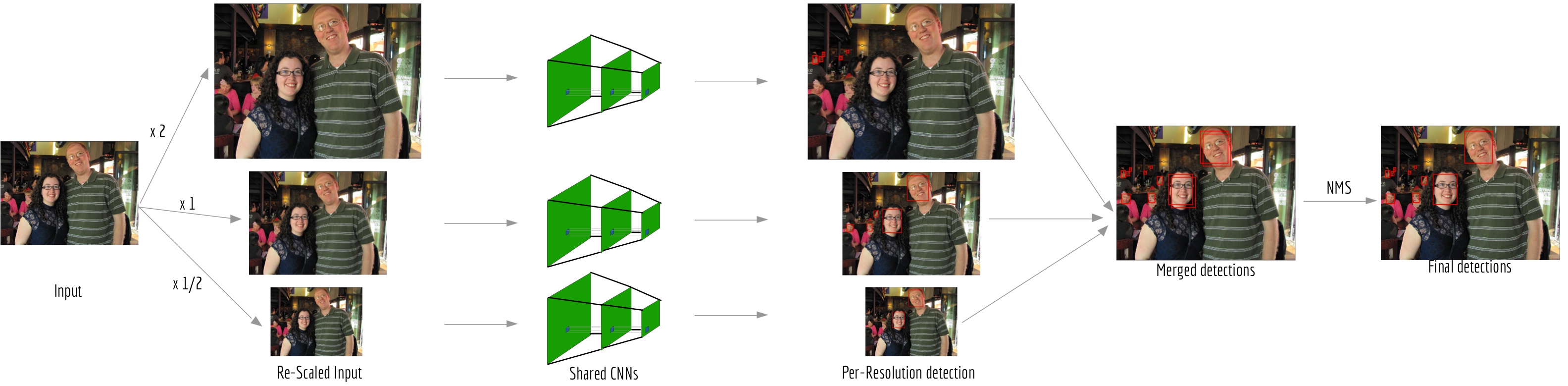} 
  \caption{Overview of the detection pipeline}
  \label{archi}
\end{figure*}

Let us now detail the detection pipeline. Starting with an input image, the input are re-scaled (multiple interpolations) and scale-specific (the scales are power of 2, depending on the input image resolution) detector are used to guarantee the scale invariance. Then, the scaled input serve as entries to a single Convolutional Neural Network (CNN) to predict response maps at every resolution. We thus extract the per-resolution detections (bounding boxes) that we merge afterwards. In the end, we apply non-maximum suppression (NMS) at the original resolution to remove the overlapped bounding boxes and obtain the final detection. We visualize the proposed architecture in Figure~\ref{archi}.
Also, after many tries, we have unfortunately not succeeded in training the model and thus, in what follows, we focus on the inference part using a pretrained model of Tiny Faces on WIDERFACE, which relies on a pretrained ResNet101[13].

\subsection{Metrics}
In order to measure the effects of each of the previously introduced key aspects, we have implemented from scratch the Jaccard similarity (\textit{Intersection-over-Union}, denoted as $J(k, k')$) that enables the distinction between true and false predicted positive bounding boxes for each ground truth. This metric also allows for the automatic matching of the predicted bounding box to their corresponding ground truth. Indeed, we consider as false positive every predicted bounding box which has a Jaccard similarity with a ground truth box lower than $0.5$. Then,  we have used the AP and implemented the ratio of true positive bounding boxes over the number of ground truth bounding boxes which is the most relevant metric for our experiments (see Appendix~\ref{sec:quant}). $$TP/GT=\frac{card(\{k \in \text{Predictions} | J(k,k_{truth})>0.5\})}{card(\{\text{Ground Truth}\})}$$

\noindent Finally, we achieved a \textbf{87\%} AP compared to the \textbf{92\%} of the original paper, on the easy validation set of WIDERFACE.

\subsection{Image resolution influence}
The performance of the Tiny Faces algorithm is linked with the image resolution. Indeed, we experimented (see Appendix~\ref{sec:quant} and~\ref{sec:qual} for qualitative and quantitative results) by downscaling an original image and plotted both the mean Jaccard similarity and the number of faces detected (Ground Truth and Predictions). We conclude that the smaller the scale, the worse the face detection with respect to the number of detected faces and AP. 

\subsection{Blurred Faces}
We also tested the influence of blur on the performance of the Tiny Faces algorithm. We thus experimented on the WIDERFACE annotated images that include heavy blurred faces (see Appendix~\ref{sec:quant}). We notice that the algorithm performs far less well in detecting blurred faces than in general.

\subsection{Benchmark}
We aim at comparing the Tiny Faces algorithm with other pretrained face detection models. In order to do so, we used Faster R-CNN[2] \href{https://github.com/tornadomeet/mxnet-face}{\textit{(customized for face detection)}}, Multi-task Cascaded Convolutional Networks [3] \textit{(MT-CNN)}, Haar Cascade[4] and Histograms of Oriented Gradients (HOG) [5]. We tested on two particular subfolders of the WIDERFACE dataset (\textit{Parade} -see Appendix~\ref{sec:qual}- and \textit{Dresses}) and we show the results we obtained in Table~\ref{tab:ben}. We can conclude that Tiny Faces outperforms all the other tested algorithms and even more when dealing with many faces.
\begin{table}[h]
\centering
\begin{tabular}{lcc|cc}
\hline
    & \multicolumn{2}{l}{Category 1} & \multicolumn{2}{l}{Category 2} \\
   Algorithms & TP/GT  & Time & TP/GT  & Time \\
  \hline
   HOG & 2\% & 9s & 54\% & 10s\\
   Haar Cascades & 8\% & 20s & 60\% & 22s\\
   Faster R-CNN & 10\% & 281s & 77\% & 178s\\
   MT-CNN & 31\% & 87s & 82\% & 64s\\
   Tiny Faces & \textbf{65\%} & 86s & \textbf{93\%} & 88s\\
\hline
\end{tabular}
\caption{Validation performance of the tested algorithms on two categories of WIDERFACE : Parade (category 1 : $\sim$ 34 faces/image) and Dresses (category 2 : $\sim$ 3 faces/image)}
\label{tab:ben}
\end{table}

\begin{figure}[h]
\centering
  \includegraphics[height=4cm, width=7cm]{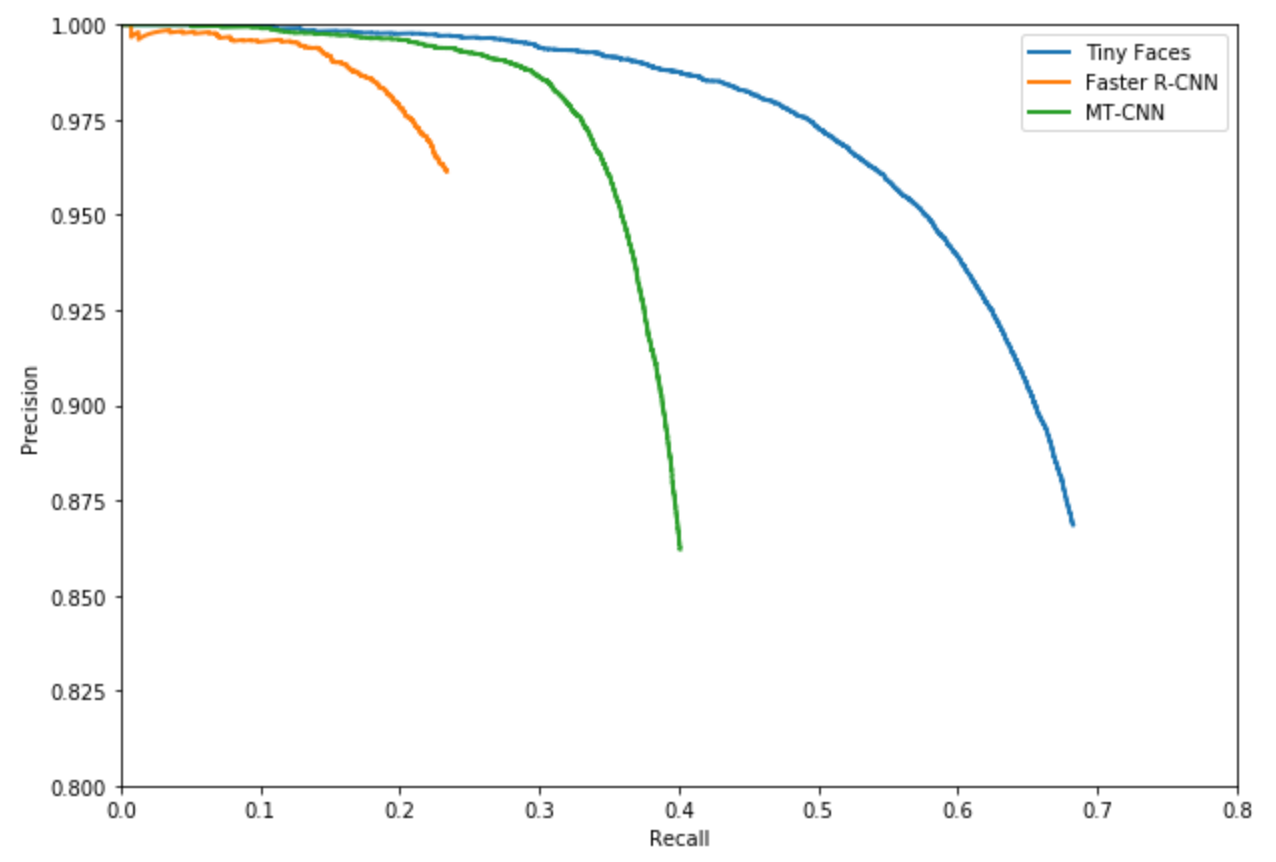}
  \caption{Precision recall curve of Tiny Faces, MT-CNN and Face Faster R-CNN on the full WIDERFACE validation set}
  \label{pr_curve}
\end{figure}

\section{Application : counting people}
Taken that counting people in a public demonstration can be a tedious task, we would like to adapt the previous approach to automate it. Thus, we aim at building a Python pipeline to detect and count people in a video of a demonstration. It would consist in detecting the faces, finding the most similar ones between frames and in the end, counting people (i.e. faces) with no duplicates.
\subsection{Dataset}
For this application, we used a music video clip \textit{(L'abreuvoir en f\^ete[6])} of a public audience partially occluded by a performer. We have selected two extracts (of around $5$ seconds each) and manually annotated these clips. Our goal is indeed to predict the number of unique people and hence, for each extract, we count the number of faces. In order to do so, we have to match people across frames, we thus had to manually match some people from different frames, two by two. \\
The frame rate in our video is $\sim25\ FPS$, so we can assume that a person does not radically move from one frame to another. However, the same person might be in a different spatial position after $10$ frames and that explains our choice of using 3 frames per second for counting people.

\begin{figure*}
\centering
  \includegraphics[width=\textwidth]{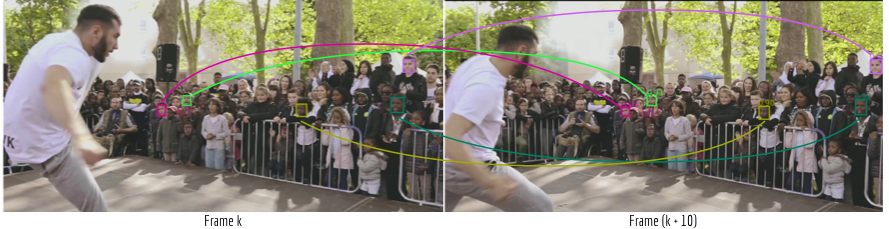}
  \caption{Illustration of the face matching between two frames using a SVM for 4 detected faces and their respective matching scores}
  \label{match}
\end{figure*}

\subsection{Face Embedding}
In order to count people only once, we have to recognize each face and then match them in the next frame. 
Obviously, we first needed to detect faces in each frame. To do so, we applied the Tiny Faces algorithm for the detection of all the faces and hence, for each frame, we got a list of predicted detected faces. Then, we have to match them across frames. We created a face embedding for each face to make the matching easier : we wrapped each picture so the faces are always in the same direction. The face alignement is achieved by a pretrained face landmarks estimation algorithm [7] and affine transformations. To overcome the fact that we do not have many images of each face and in order to have a cheaper (in terms of time and memory) algorithm comparison, we used a pretrained CNN for face embedding [8] (generating $128$ basic measurements for each face). 

\subsection{Matching}
Now that we have an embedding for each face, we can match the same faces across frames. We would do it in a one-vs-all manner, i.e. build a model that can predict the most similar faces from one frame to the other using a simple classifier. To train the previous, we first needed to augment our dataset. As a person does not radically move from one frame to one right after, we can use the predicted bounding box coordinates for a face in the next two frames. Then, we made use of data augmentation techniques such as adding gaussian noise and also random values to each RGB channel, using \href{https://github.com/aleju/imgaug}{\textit{imgaug}} Python library. Finally, we obtained $10$ different images per face, that served as the positive samples of the training set. As for the negative ones, we randomly chose $10$ other faces. After training multiple binary classifiers (one for each face) per frame, we predicted the most similar face in the same geographical neighbourhood (a $600px$ square box, we assume that the camera and the people in its visual field do not move really fast) of another frame. The classifiers used are linear SVMs. \\Eventually, we have the most similar faces in another frame as shown in Figure~\ref{match}. However, we have to fix a threshold with respect to the SVM distance in order to remove false positives. This threshold was calculated using cross-validation to maximize the $F_{0.5}$-score (more weights to precision) and the annotated matched faces. 
\subsection{Counting}
Finally, using all the previous, we are able to build an incremental count of the number of unique faces as the the video is moving forward. Indeed, for each analyzed (1/10) frame, we count the number of detected people and remove, in the incremental count, the number of already seen ones (using the face matching). The number of unique people is simply the sum of new people in each frame, paying attention to count them only once across all frames.\\
The results are displayed in the table below : \\

\begin{tabular}{cccc}
\hline
  Clip & Ground Truth  & Prediction & AP\\
  \hline
   1 & 139 & 141 & $98\%$\\
   2 & 148 & 156 & $95\%$\\
\hline
\end{tabular}

\section{Extensions}
We have seen that \textit{Tiny Faces} outperforms some of the recent face detection algorithms when it was released (04/2017). 
However, in future work, we would like to compare it with the most recent and accurate face detection algorithms e.g FaceBoxes[9] (better on \href{http://vis-www.cs.umass.edu/fddb/results.html}{\textit{FDDB}} , Appendix A), Single Stage Headless Face Detector[10] or RetinaNet[11].
As explained above, we didn't succeed at training the algorithm. Thus, we would like to try again and train on novel pretrained models e.g. ResNeXt[12] for better accuracy.\\
Also, the general approach was to detect tiny objects in images but we only focus (as the original paper) on faces but we think that we could apply this approach to other small object such calcifications in mammography pictures. \\
Moreover, we used a simple pipeline and approach to track and count people across frames. We let, for future work, the combining with tracking algorithms such as DeepSort[13] .

\section{Conclusion}
To conclude, let us recall that we performed an in-depth analysis of this novel and interesting approach to understand the key parts before successfully applying it to the counting of people in a public demonstration. We were able to compare tiny faces with state-of-the-art (04/2017) face detection algorithms and concluded that it outperforms the rest.

\section*{References}\fontsize{7.5}{1.2pt}\selectfont
\noindent[1] Hu and Ramanan. Finding tiny faces. CoRR, abs/1612.04402, 2016.

\noindent[2] Ren, He, Girshick, and Sun. Faster R-CNN towards real-time object detection with region proposal networks. CoRR, abs/1506.01497, 2015.

\noindent[3] K. Zhang et al. Joint face detection and alignment using multi-task cascaded convolutional networks. CoRR, abs/1604.02878, 2016. 

\noindent[4] Viola and Jones.  Rapid object detection using a boosted cascade of simple features, 2001.

\noindent[5] Navneet Dalal and Bill Triggs. Histograms of oriented gradients for human detection. In CVPR, 2005.

\noindent[6] FAMVK EVENTS: L'abreuvoir en f\^ete. http://bit.ly/2BcPBaE. 

\noindent[7] Kazemi and Sullivan. One millisecond face alignment with an ensemble of regression trees. In CVPR. IEEE Computer Society, 2014.

\noindent[8] Schroff, Kalenichenko, and Philbin. Facenet: A unified embedding for face recognition and clustering. CoRR, abs/1503.03832, 2015.

\noindent[9] Zhang, Zhu, Lei, Shi, Wang, and Li. Faceboxes: A CPU real-time face detector with high accuracy. CoRR , abs/1708.05234, 2017.

\noindent[10] Najibi, Samangouei, Chellappa, and Davis. SSH: single stage headless face detector. CoRR, abs/1708.03979, 2017.

\noindent [11]   Lin,  Goyal,  Girshick,  He,  and Doll\'ar. Focal loss for dense object detection. CoRR, abs/1708.02002, 2017.

\noindent[12] Xie, Girshick, Doll\'ar, Tu, and He. Aggregated residual transformations for deep neural networks. CoRR, abs/1611.05431, 2016.

\noindent[13] Wojke, Bewley, and Paulus. Simple online and realtime tracking with a deep association metric. In 2017 IEEE ICIP

{
\bibliographystyle{unsrt}
\bibliography{bibli}
}

\cleardoublepage
\appendix
\section*{Appendix}
\section{Quantitative Results}\label{sec:quant}
\begin{figure}[h]
  \centering
  \includegraphics[height=5cm, width=8cm]{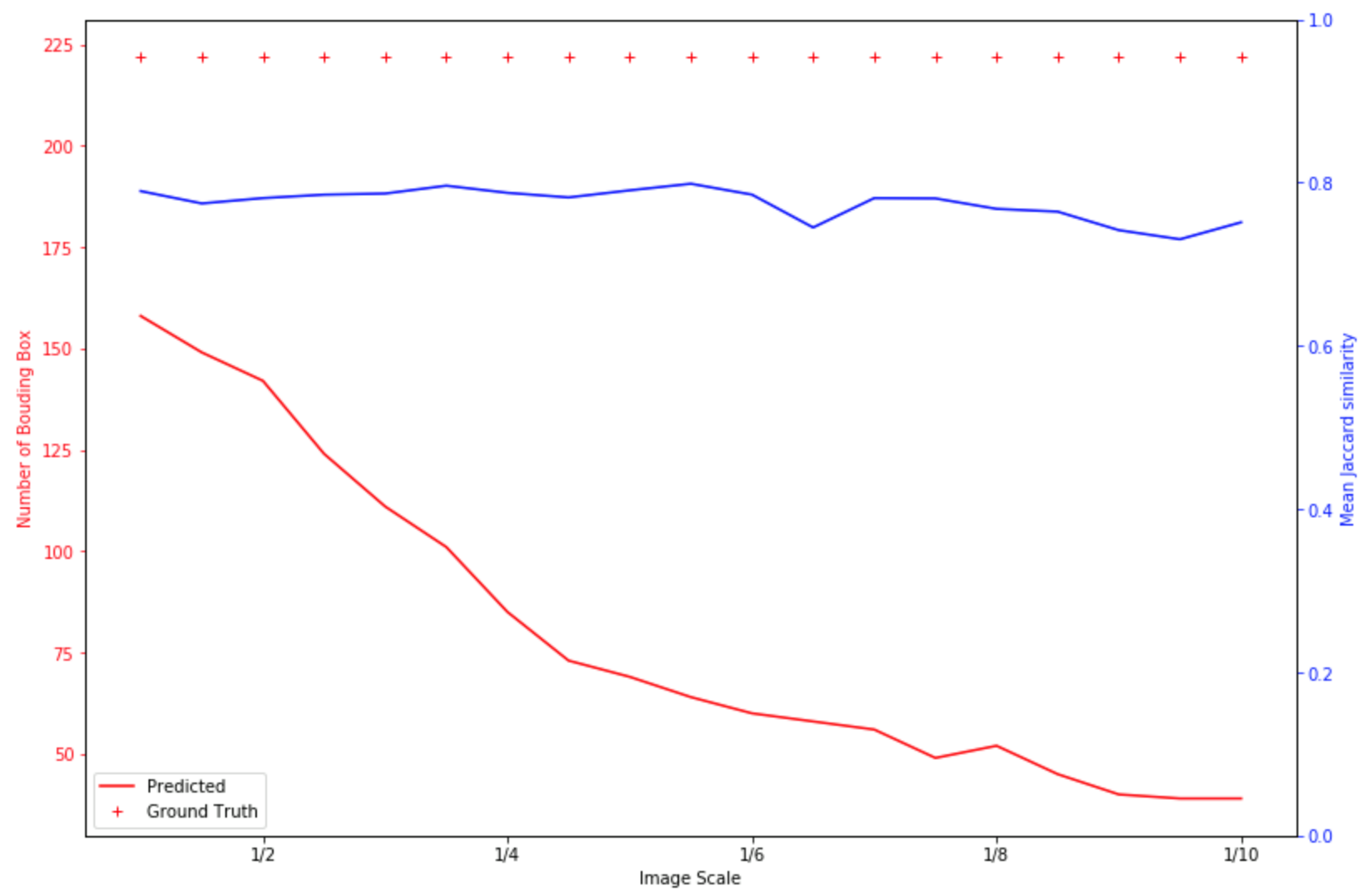}
  \caption{Ratio of true positive predicted bounding boxes over ground truth boxes and Jaccard similarity on the WIDERFACE Parade category when downscaling input images.}
\end{figure}

\begin{figure}[h]
  \centering
  \includegraphics[height=5cm, width=8cm]{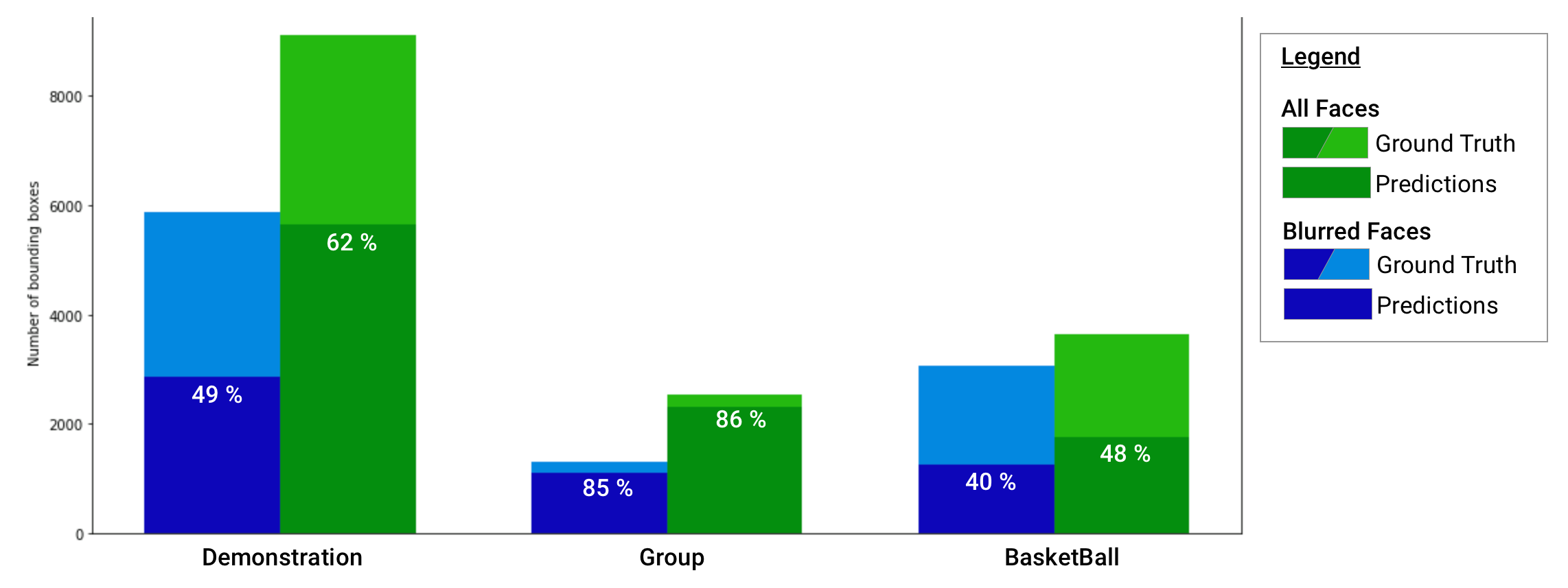}
  \caption{Performance of the Tiny Faces algorithm when dealing with heavy blurred faces (as annotated in the WIDERFACE set).}
\end{figure}

\begin{figure}[h]
  \centering
  \includegraphics[width=8cm]{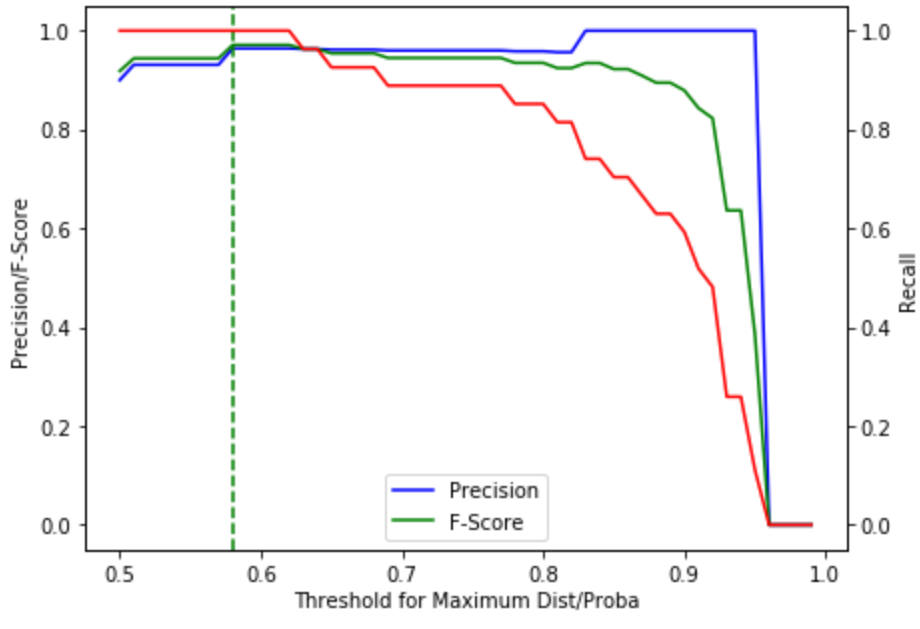}
  \caption{Precision, Recall and $F_{0.5}$ curves to determine the threshold for the SVM matching classifier.}
\end{figure}

\begin{figure}[h]
  \centering
  \includegraphics[height=4cm, width=8cm]{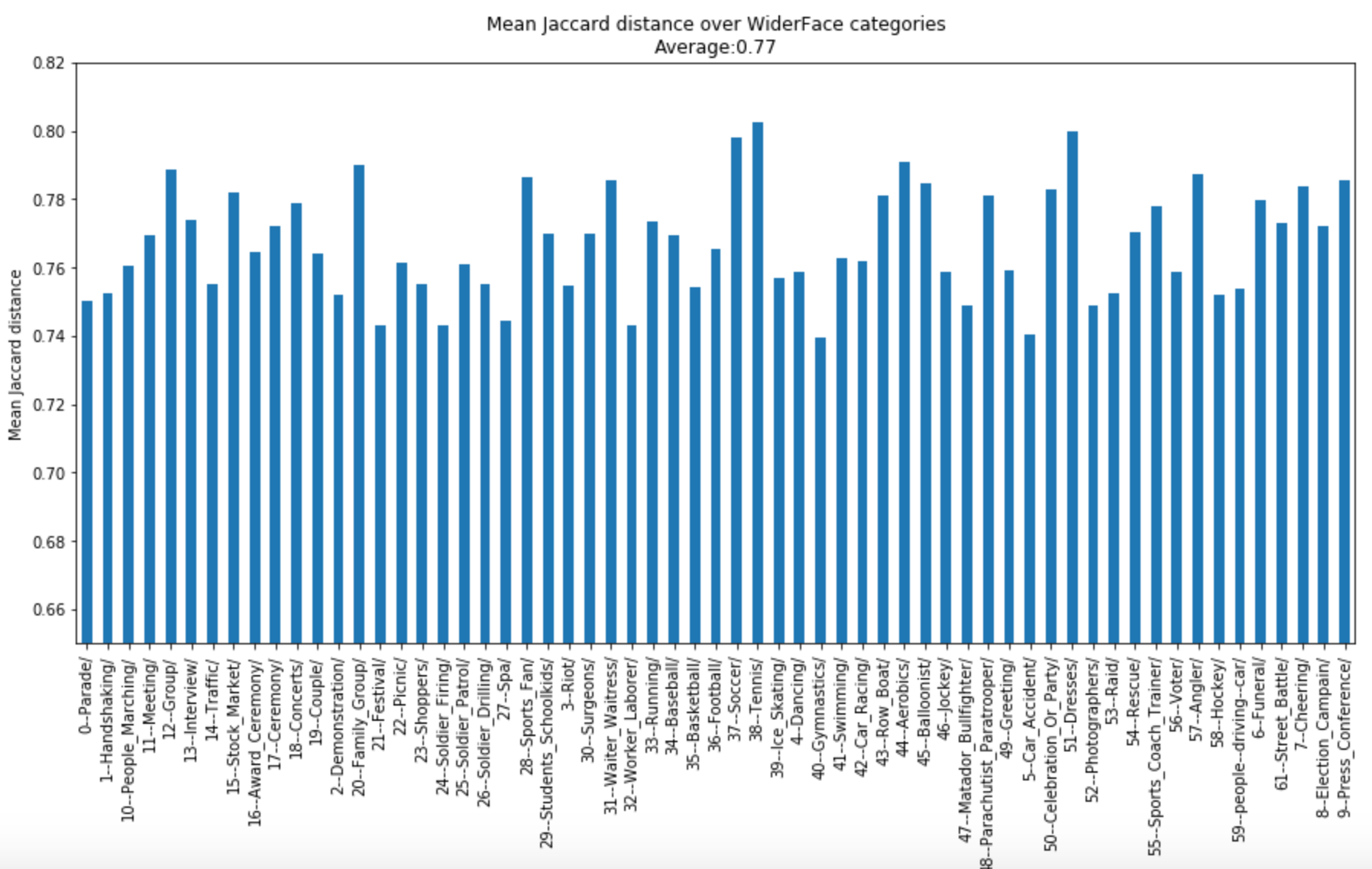}
  \caption{Mean Jaccard on the validation set of WIDERFACE. Average : $77\%$.}
\end{figure}

\begin{figure}[h]
  \centering
  \includegraphics[height=4cm, width=8cm]{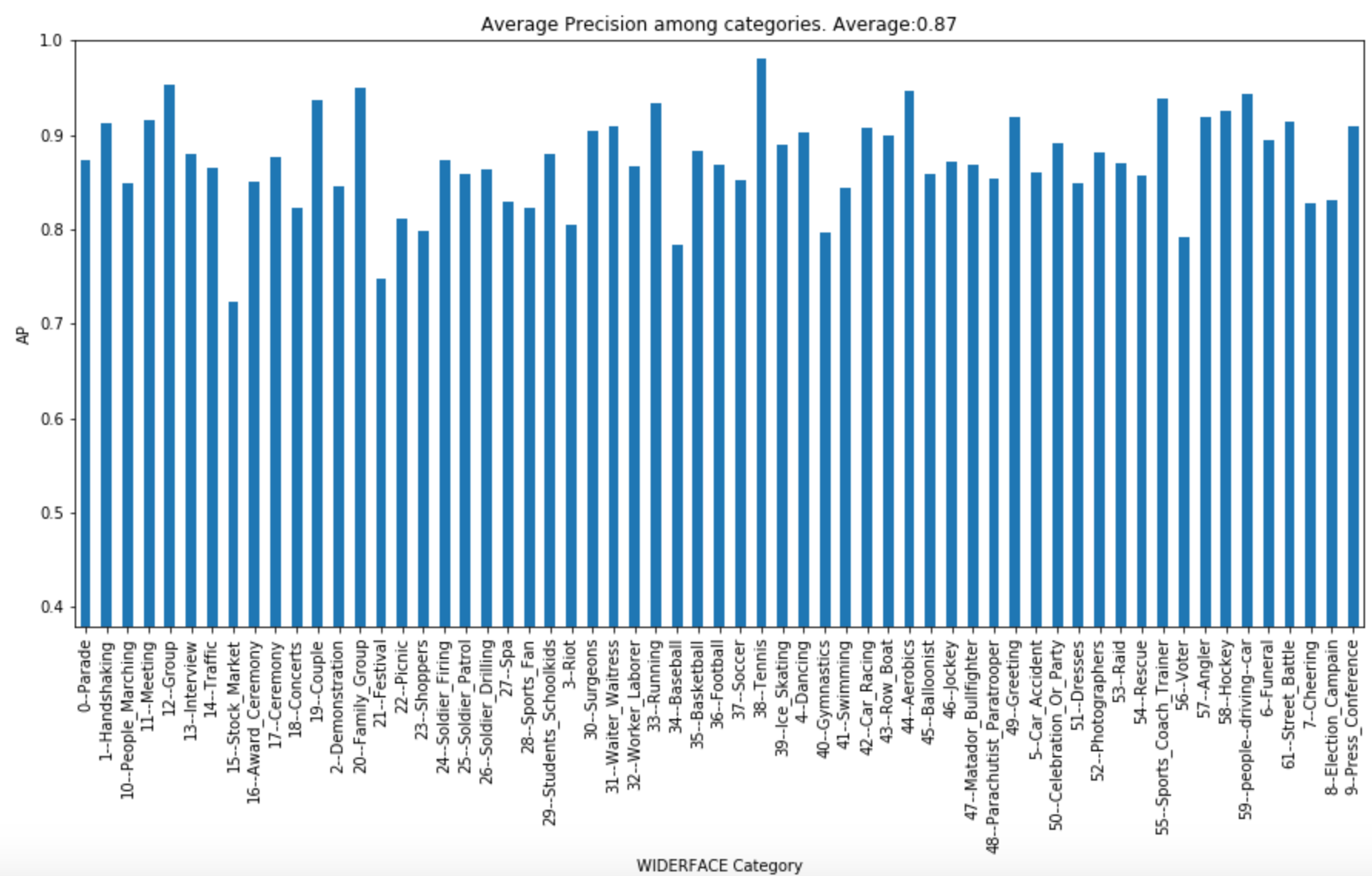}
  \caption{AP Performance on the validation set of WIDERFACE (No effect of the missed GT boxes on AP). Average : $87\%$.}
\end{figure}

\begin{figure}[h]
  \centering
  \includegraphics[height=4cm, width=8cm]{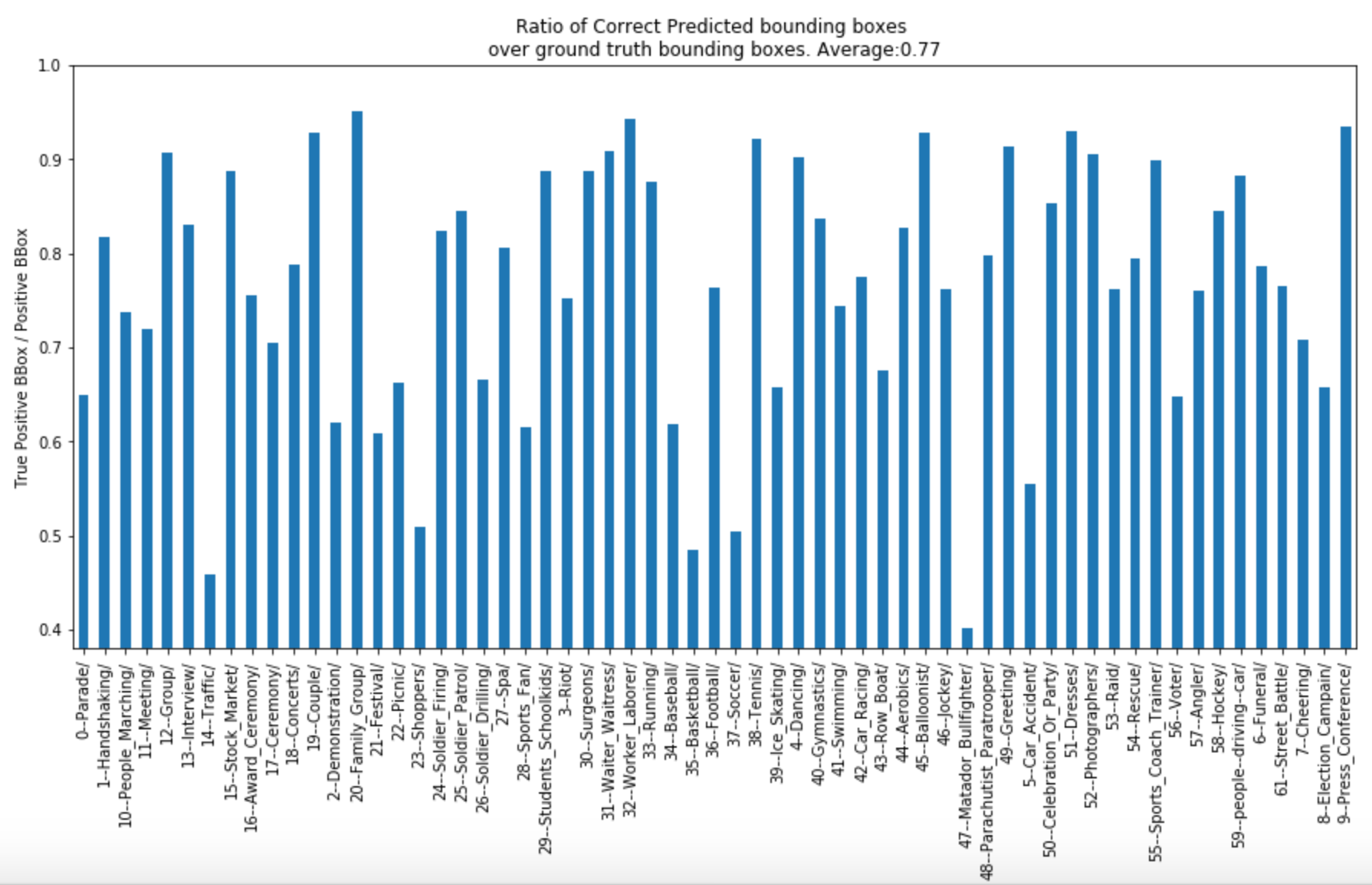}
  \caption{Unbalanced ratio of true positive predicted bounding boxes over ground truth boxes of Tiny Faces. Average : $77\%$.}
\end{figure}

\begin{figure}[h]
  \centering
  \includegraphics[height=5cm, width=8cm]{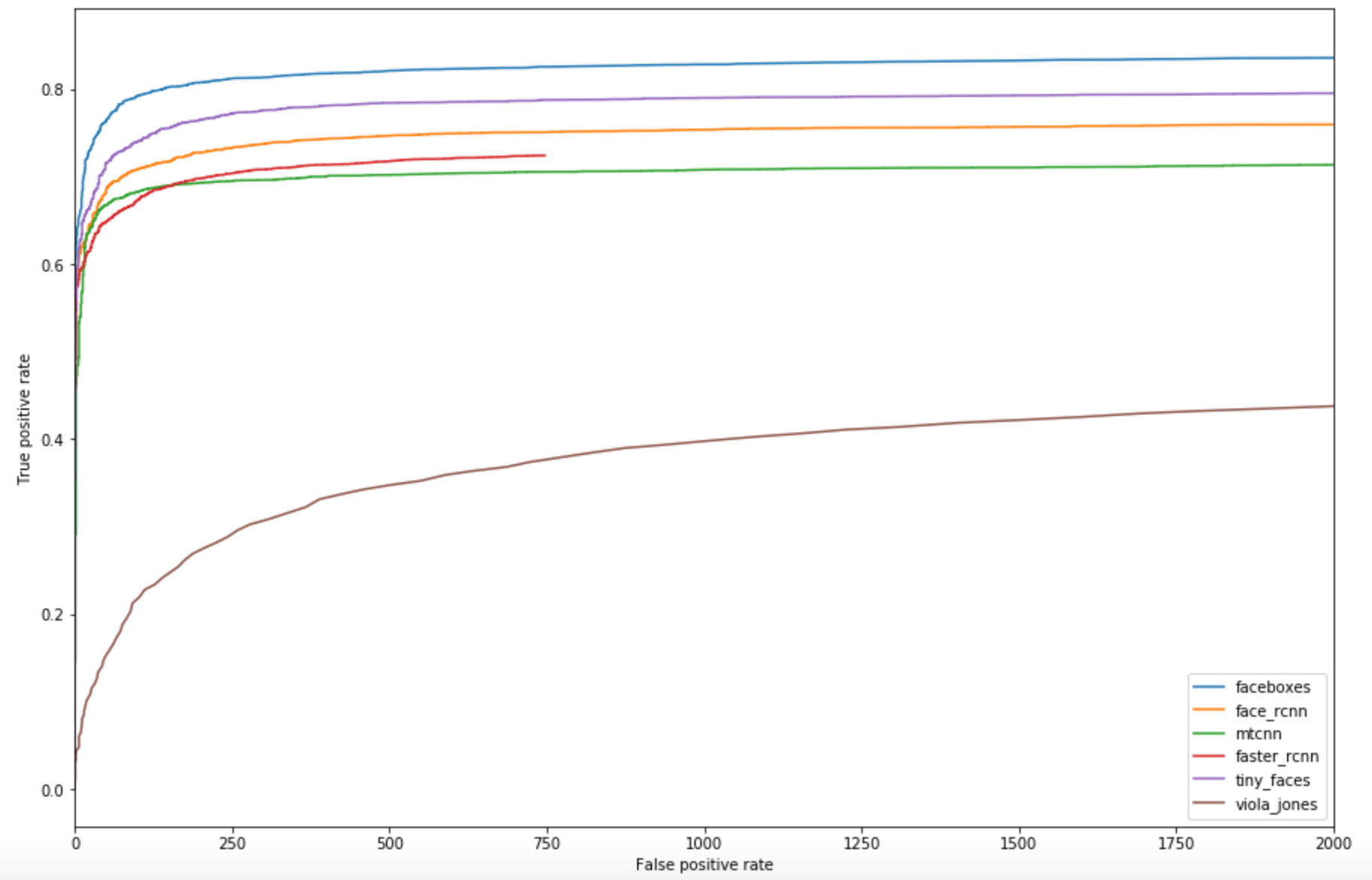}
  \caption{ROC curves generated using \href{http://vis-www.cs.umass.edu/fddb/results.html}{FDDB results} of some of the best face detection algorithms. Tiny Faces performs well but best performance achieved with \textit{FaceBoxes}.}
\end{figure}

\cleardoublepage
\section{Qualitative Results}\label{sec:qual}

\begin{figure}[h]
  \centering
  \includegraphics[width=8cm]{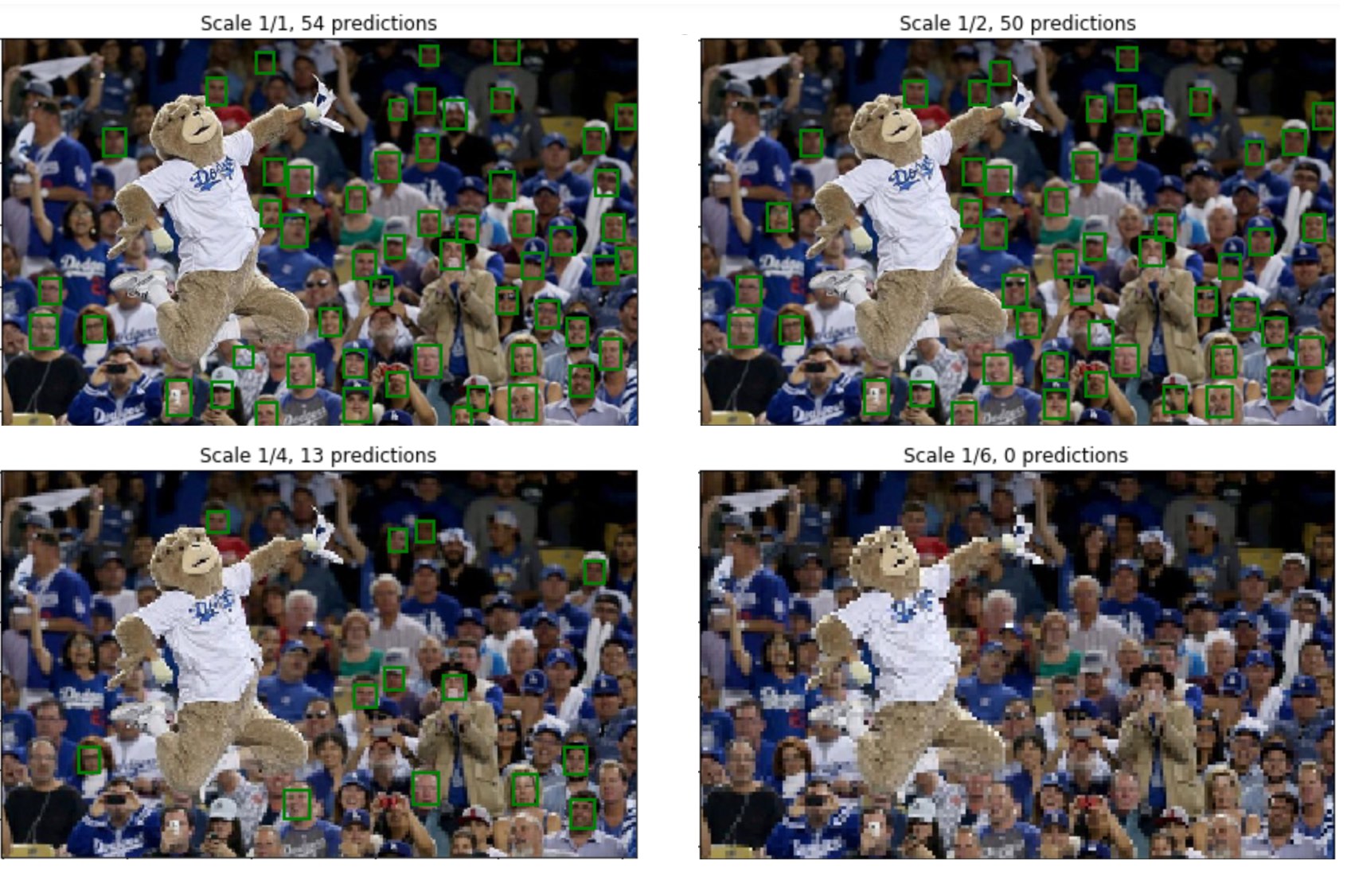}
  \caption{Influence of image resolution over the number of face detected on the same input image downscaled multiple times.}
\end{figure}

\begin{figure}[h]
  \centering
  \includegraphics[height=5cm, width=8cm]{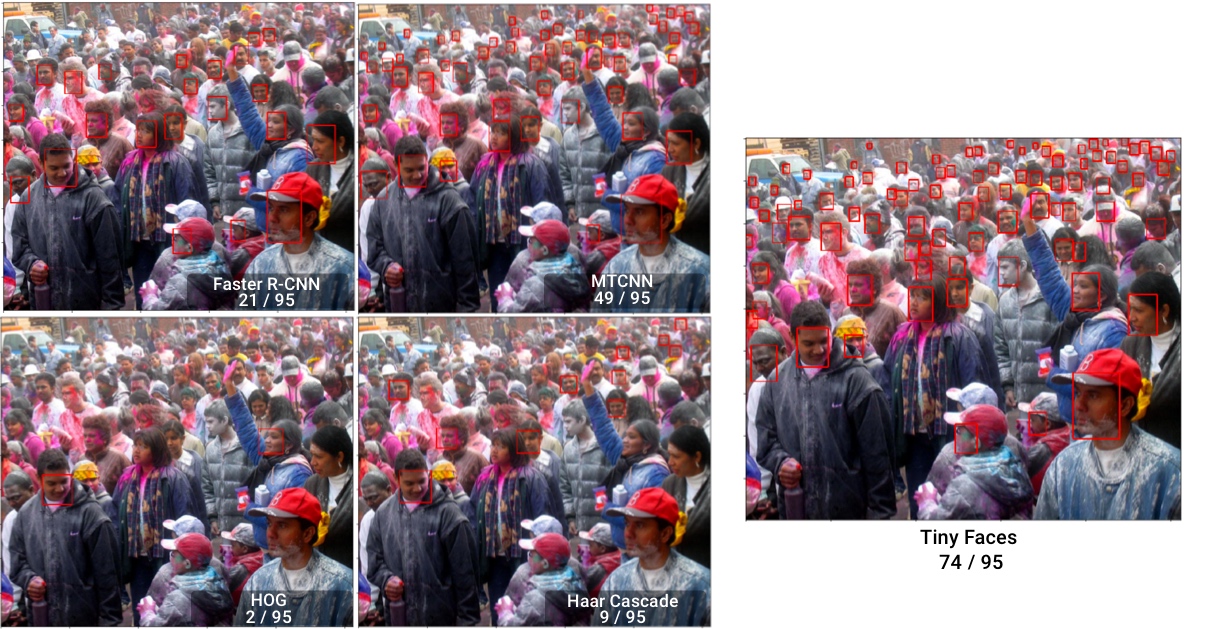}
  \caption{Comparison of multiple face detection algorithms and Tiny Faces on a particular image of the WIDERFACE Parade category. TinyFaces outperforms the other algorithms.}
\end{figure}

\begin{figure}[h]
  \centering
  \includegraphics[height=5cm]{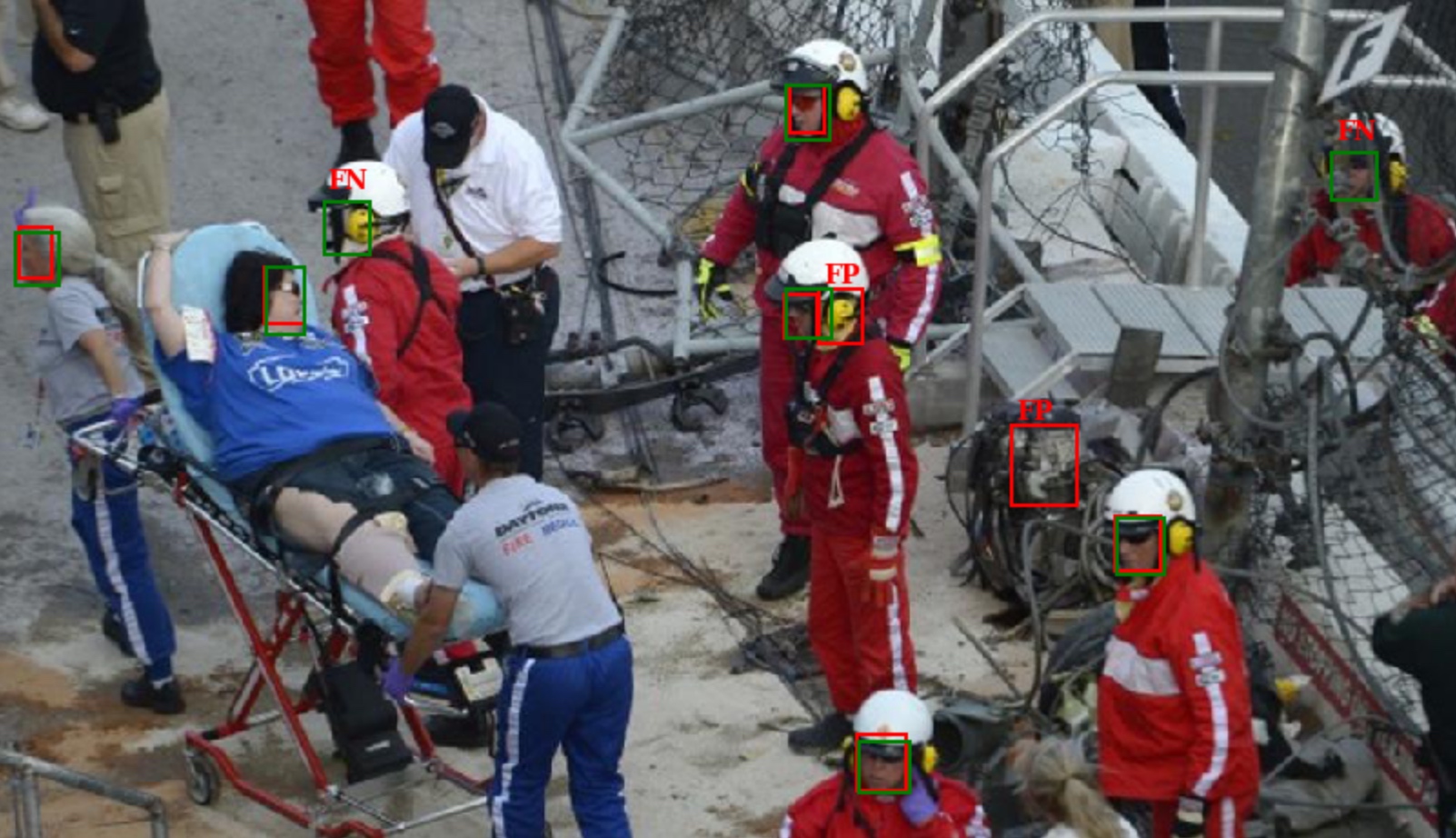}
  \caption{Examples of rare false positive and false negative predictions. Ground truth bounding boxes are plotted in green and predicted bounding boxes are plotted in red. We can spot $2$ false positives and $2$ false negatives.}
\end{figure}

\begin{figure}[hbt]
\centering
\includegraphics[height=5cm]{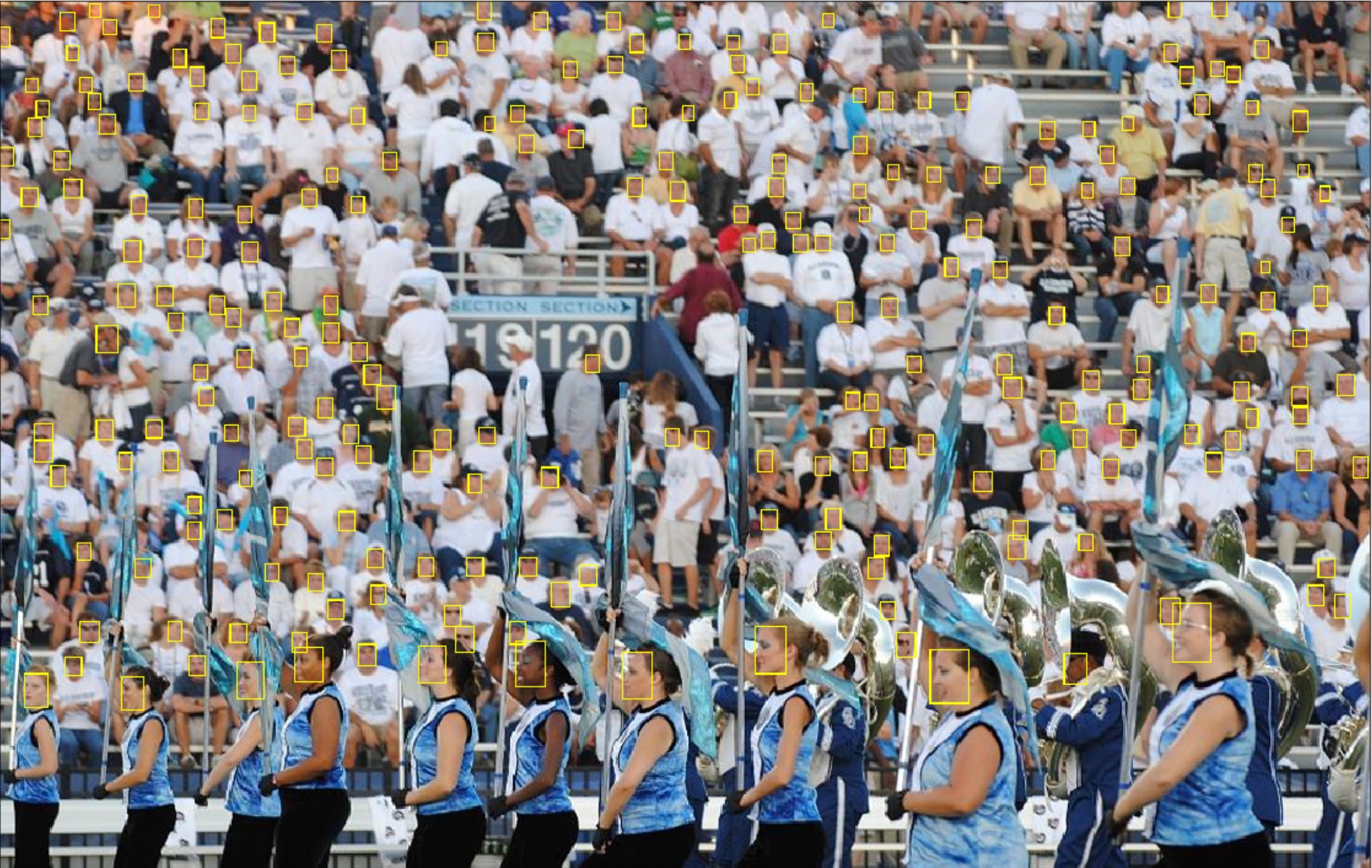}
  \caption{Examples of Tiny Faces impressive results : 231 predicted detections over the 283 ground truth bounding boxes on one image of the WIDERFACE \textit{Parade} category.}
\vspace{1cm}
\includegraphics[height=5cm]{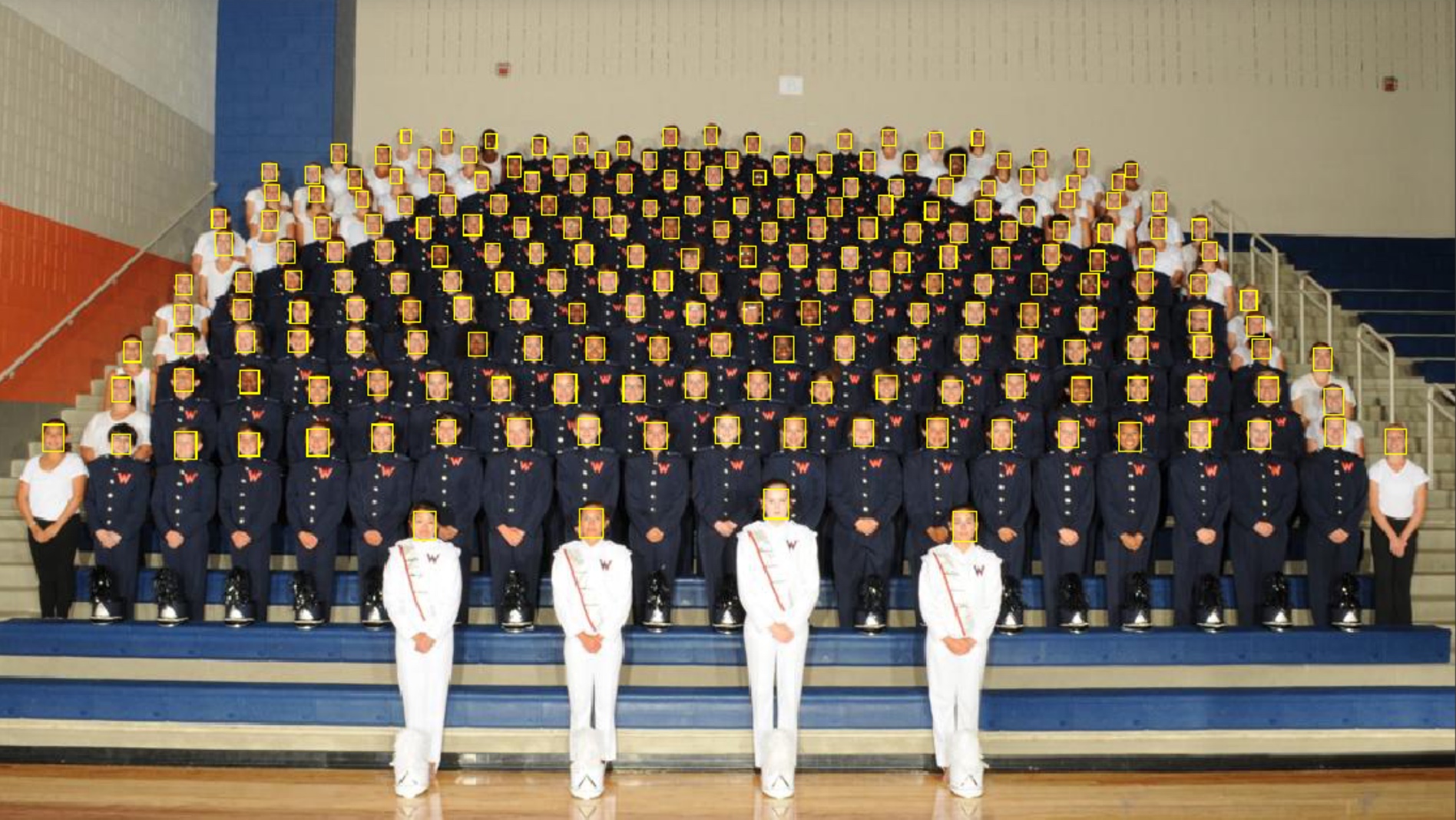}
  \caption{Examples of Tiny Faces impressive results : 221 predicted detections over the 221 ground truth bounding boxes on one image of the WIDERFACE \textit{Parade} category.}
\label{fig:another}
\end{figure}
\end{document}